%% file: sample-sigplan.tex
\documentclass[sigplan,screen]{acmart}
\AtBeginDocument{%
  }

\usepackage{booktabs}
\usepackage{multirow}
\usepackage{makecell}
\usepackage{array}
\usepackage{xcolor}
\usepackage{etoolbox}
\makeatletter
\patchcmd{\authornote}{\g@addto@macro\addresses{\@authornotemark}}{}{}{}
\makeatother
\renewcommand\footnotetextcopyrightpermission[1]{}
\setcopyright{none}

\begin{document}

\title{The Agent Behavior: Model, Governance and Challenges in the AI Digital Age}

\author{Qiang Zhang$^{1 \dagger}$,Pei Yan$^{3 \dagger}$, Yijia Xu$^{1,2 \ast}$, Chuanpo Fu$^{4}$, Yong Fang$^{1}$, Yang Liu$^{2}$}

\authornote{ The Yijia Xu is the corresponding author (email: xuyijia@scu.edu.cn).}
\authornote{ The Qiang Zhang and Pei Yan contributed equally to this research.}

\affiliation{%
	\institution{$^{1}$School of Cyber Science and Engineering, Sichuan University, China}
	\country{}
}
\affiliation{%
	\institution{$^{2}$College of Computing and Data Science, Nanyang Technological University, Sinapore}
	\country{}
}
\affiliation{%
	\institution{$^{3}$College of Electronics and Information Engineering, Shenzhen University, China}
	\country{}
}
\affiliation{%
	\institution{$^{4}$Department of Computer Science and Technology, Tsinghua University, China}
	\country{}
}


%

\renewcommand{\shortauthors}{Zhang et al.}

\begin{abstract}
Advancements in AI have led to agents in networked environments increasingly mirroring human behavior, thereby blurring the boundary between artificial and human actors in specific contexts. This shift brings about significant challenges in trust, responsibility, ethics, security and etc. The difficulty in supervising of agent behaviors may lead to issues such as data contamination and unclear accountability. To address these challenges, this paper proposes the "Network Behavior Lifecycle" model, which divides network behavior into 6 stages and systematically analyzes the behavioral differences between humans and agents at each stage. Based on these insights, the paper further introduces the "Agent for Agent (A4A)" paradigm and the "Human-Agent Behavioral Disparity (HABD)" model, which examine the fundamental distinctions between human and agent behaviors across 5 dimensions: decision mechanism, execution efficiency, intention-behavior consistency, behavioral inertia, and irrational patterns. The effectiveness of the model is verified through real-world cases such as red team penetration and blue team defense. Finally, the paper discusses future research directions in dynamic cognitive governance architecture, behavioral disparity quantification, and meta-governance protocol stacks, aiming to provide a theoretical foundation and technical roadmap for secure and trustworthy human-agent collaboration.
\end{abstract}

\keywords{Agent, Behavior Governance, Cyber Confrontation, Large Language Model}


\maketitle

\section{Introduction}

The Agent refers to an entity capable of perceiving the environment, autonomously making decisions, and executing actions to achieve a goal~\cite{xi2025rise,xie2024large,durante2024agent}. With the rapid development of artificial intelligence technologies, agents are approaching human intelligence levels and mimicking human behavior patterns. Moreover, agents can absorb expert thinking, models, and knowledge, thereby replacing human work in many tasks. In the future, agents will become increasingly human-like, making it nearly impossible to determine in a black-box environment whether a task, job, or decision was completed by a human or an agent~\cite{wang2024ali,xie2024can,sreedhar2024simulating,song2022will,cheng2022human}. Such a development trend could entail unforeseen challenges with potentially profound implications for human society. Therefore, referencing the understanding of human behavior, we propose the concept of "agent behavior", and conduct deeper research into the agent and human behavior.
Specifically, the behavior mentioned in this article refers to network behavior for better description and discussion.

We model agent behavior with reference to human behavioral science. Human behavior is only a surface phenomenon, which is influenced by underlying psychological activities and brain cognition. When a person is stimulated by the external environment, the stimulus is transmitted to the brain through sensory receptors. The brain then activates the neural system and generates a series of psychological activities, and human behavior is the objective manifestation of these psychological activities in the real world. Compared to humans, an agent's behavior is not influenced by psychological activities or brain cognition, but the process of behavior generation can be analogized to that of humans. From the agent’s perspective, external prompts are processed by computer hardware; the processed information is then fed into a large language model for in-depth reasoning and deduction. The results of this reasoning and deduction are then used to perform specific actions in the real or virtual world through API interfaces.

\begin{figure}[h]
	\centering
	\includegraphics[width=0.457\textwidth]{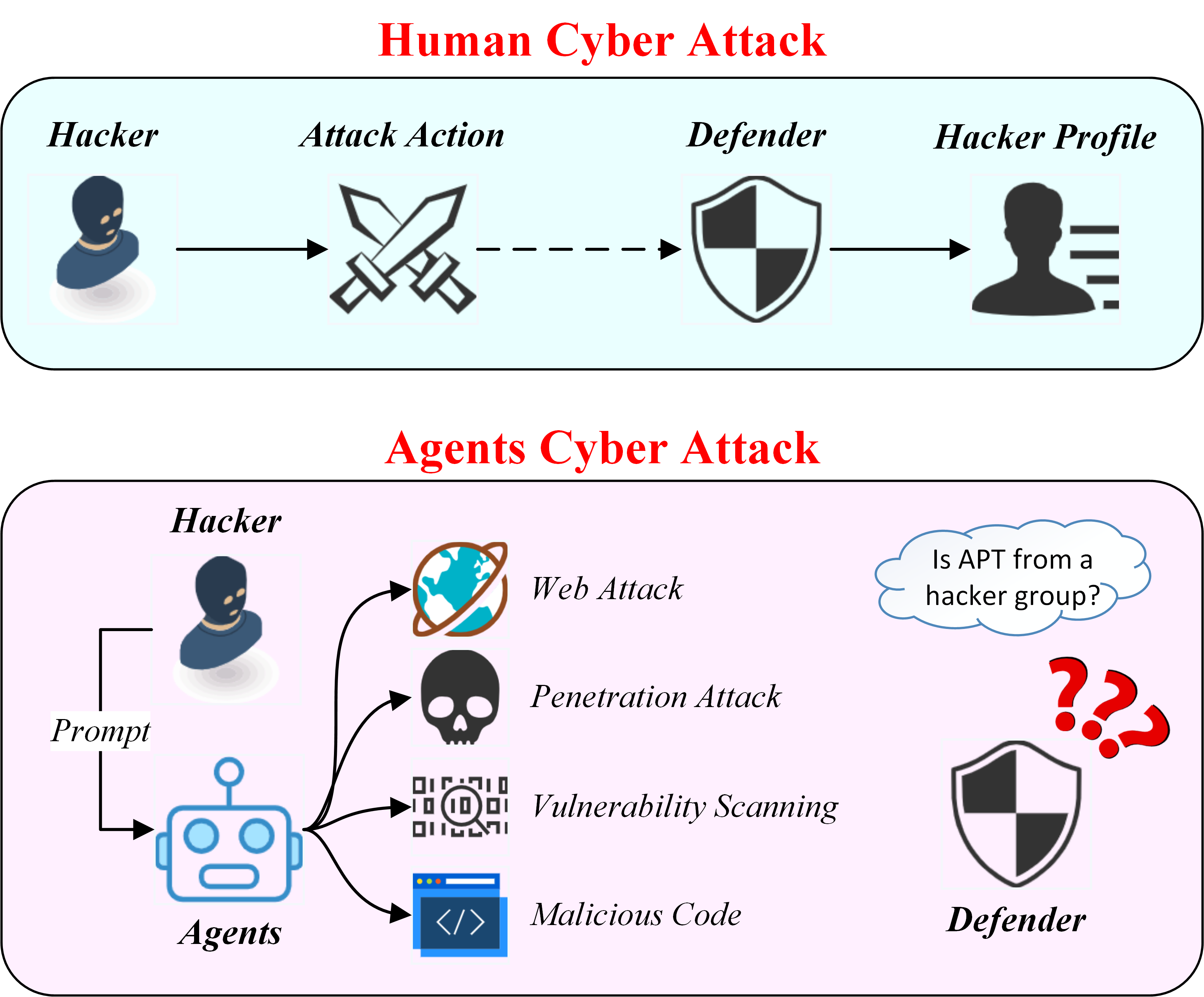}
	\caption{The comparison of cyber attack procedure between human and agents.
		When humans carry out cyber attacks, defenders will profile the attackers based on the attacks, and use this profiling information for tracing the source and enhancing their defense capabilities.
		If humans use Agents to conduct cyber attacks, the agents can perform more diverse attack behaviors and mislead the defenders' tracking and profiling directions.}
	\label{fig1}
\end{figure}

As agents become increasingly similar to humans, the need to regulate and govern their behavior is also growing, given the profound implications for trust, accountability, and ethical considerations. In the field of cybersecurity, attackers may use intelligent agents to execute automated attacks and carry out Advanced Persistent Threats~(shown in Figure \ref{fig1}), making it difficult to create profiles for them~\cite{mamoon2025adaptive,du2025cyber}. In recommendation systems, intelligent agents can quickly modify or reset user profiles, complicating the capture of accurate user information~\cite{zhang2024generative}. Intelligent agents may generate simulated data that diverges from real data, affecting the authenticity of data collected by research institutions and leading to data contamination that can destroy baseline models~\cite{green2018data,blythe2024synthetic,sengupta2024mag}. In online psychological therapy, if the psychologist is an intelligent agent, its therapeutic effectiveness may not match that of humans, and negative emotions from patients could damage the agent's functionality. In online medical consultations, interactions with intelligent agents instead of real doctors may raise trust issues, with high error risks and challenges in addressing liability~\cite{chen2025self,farber2024physicians}. These potential scenarios represent only the tip of the iceberg. If agent technologies are not deployed under controlled conditions, their negative impacts are likely to permeate every aspect of daily life in the AI-dominated digital era.

Nevertheless, in advancing toward the governance of agent behavior, identifying effective means to distinguish agent behaviors from human behaviors amid massive behavioral datasets constitutes the foundational step. After rigorous evaluation, we believe that using specialized agents to complete behavioral judgment tasks is a relatively feasible method \cite{talebirad2023multi,gladden2022empowerment}. Agents excel at processing large volumes of data and performing complex pattern recognition, allowing them to quickly analyze and identify behavioral characteristics. They can continuously learn and adapt to new behavior patterns, improving their recognition accuracy by calling tools or algorithms~\cite{hussain2024tutorial}. For example, in recommendation systems, a governance agent can distinguish behavioral subjects by behavioral compliance to limit the potential adverse effects caused by agent activities, thereby enhancing the accuracy and personalization of recommendations for real users.
Therefore, this paper explores the commonalities and differences between human network behavior and agent network behavior, while analyzing the specific actions at each stage of network behavior, which we summarized into six main action phases. We defined these 6 phases as the Network Behavior Lifecycle. Based on the differences between the behavioral cycles of humans and agents, we proposed a preliminary technical approach to distinguish between the two types of behavior. Additionally, we conducted case studies to validate the feasibility of these technical approaches to a certain degree.

In light of these considerations, the evolution of agents represents not just a technological advancement, but a profound shift in how we conceptualize and interact with digital entities in various domains. As agents increasingly mirror human capabilities, they offer the potential to revolutionize industries by enhancing efficiency and reshaping traditional roles. However, this transformation also brings challenges that require careful consideration, particularly in terms of cybersecurity, data integrity, and social trust. By addressing these issues, we can harness the full potential of intelligent agents to improve human life while safeguarding against potential risks. This paper aims to explore these dynamics, beginning with an examination of the technological foundations that enable agent evolution. We then delve into the implications of agent integration across key sectors, followed by an analysis of the challenges and opportunities presented by this paradigm shift. Ultimately, we seek to articulate a vision for a future where human and agent collaboration drives innovation and societal progress, while maintaining ethical standards and trust.

\section{Background}

The study of network behavior has become an important field for understanding and optimizing human-computer interaction. The behaviors of humans and agents on the internet are similar, but there are certain differences in their behavior patterns. Before delving into these differences, it is essential to first clarify the definitions of these two types of behavior, which provides the conceptual foundation for our further in-depth discussion: 
I) \emph{Agent Behavior} is the pattern of actions exhibited by artificial intelligence systems, software, or robots within a specific environment. These behaviors are typically generated based on pre-set rules, algorithms, or learning models. 
II) \emph{Human Behavior} is the way people react and act in different environments and situations. This behavior is influenced by a variety of factors, including biology, psychology, sociocultural background, and personal experiences.
Defining these two categories of behavior fundamentally characterizes their nature, guides the exploration of directions in interpreting human network behavioral science, and establishes a theoretical basis for modeling the network behavior lifecycle.

\begin{figure*}[htbp]
	\centering
	\includegraphics[width=0.9\textwidth]{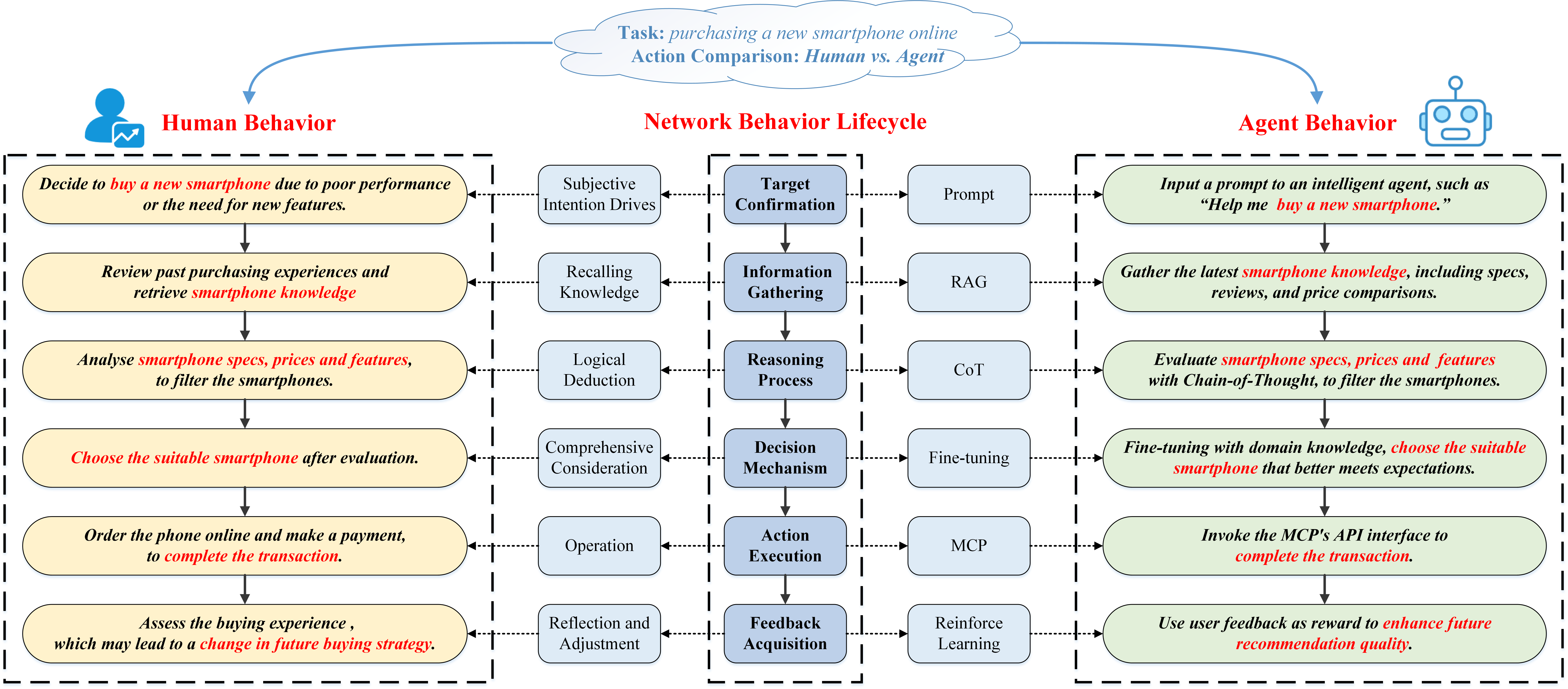}
	\caption{Using \textit{purchasing a new smartphone online} as an example, illustrates the comparison between human behavior and agent behavior within the same network behavior lifecycle.}
	\label{fig2}
\end{figure*}

\subsection{Human Network Behavioral Science}

Human network behavioral science focuses on the study of how people perceive, decide, and act within digital environments, particularly in the context of the internet and networked systems~\cite{jin2013understanding}. Unlike purely offline behaviors, human actions online are shaped by the unique affordances and constraints of digital platforms—such as instantaneous information access, persistent connectivity, and algorithm-driven content exposure. These factors not only amplify individual cognitive and emotional responses but also foster new patterns of interaction, collaboration, and even conflict in virtual spaces. As a result, human network behavior exhibits both the complexity of traditional social dynamics and the novel challenges introduced by the digital medium.

Human users display a rich tapestry of behavioral signatures, including variability in response times, susceptibility to cognitive biases, and context-dependent decision-making that often reflects social influences and emotional states. For example, the social proof principle from social psychology suggests that individuals are more likely to adopt behaviors or opinions endorsed by others, which is frequently observed in phenomena such as online reviews or viral trends~\cite{fenko2017social}. Similarly, the herding effect and bandwagon effect describe how users may follow the actions of the majority, leading to collective behaviors that can rapidly amplify within digital communities.
When engaging in online shopping, humans may be swayed by peer reviews, impulsive emotions, or previous experiences, resulting in behaviors that are nonlinear and sometimes irrational~\cite{ali2021consumer}. These nuanced patterns stand in contrast to the efficiency-driven, rule-based, and highly consistent behaviors typically exhibited by agents. Furthermore, the bounded rationality principle highlights that human decision-making is limited by available information, cognitive capacity, and time constraints, which are especially pronounced in fast-paced online environments~\cite{earl2016bounded}. By systematically studying the characteristics and lifecycle of human network behavior, researchers can better identify the subtle markers that differentiate genuine human activity from algorithmic or automated actions.

Moreover, as intelligent agents become increasingly adept at mimicking human actions, the field of human network behavioral science provides essential theoretical and methodological foundations for developing effective detection and governance strategies. Insights from behavioral economics and computational social science—such as how humans process information, adapt to feedback, and interact socially online—inform the design of systems that can more accurately model, support, and protect human users. Ultimately, a deep understanding of human network behavior not only enhances user experience and trust but also underpins the technical approaches—such as the Network Behavior Lifecycle and the Agent for Agent (A4A) paradigm—proposed in this paper for distinguishing and managing agent behaviors in complex digital ecosystems.

\subsection{Network Behavior Lifecycle}

Despite sharing surface-level similarities, the behavioral processes underlying human and agent actions are fundamentally different—shaped by distinct mechanisms of cognition, decision-making, and learning. Human behaviors are inherently subjective, influenced by emotions, social context, and reflective thought. In contrast, agents operate through algorithmic logic, optimizing actions based on structured data and feedback. To this end, we propose a network behavior lifecycle model that delineates the sequential stages through which both human and agent behaviors unfold.


The network lifecycle we proposed consists of six stages: Target Confirmation, Information Gathering, Reasoning Process, Decision Mechanism, Action Execution, and Feedback Acquisition, as illustrated in Figure~\ref{fig2}. Humans and agents exhibit distinct differences in their performance across these 6 stages. Taking "purchasing a smartphone online" as an example, humans often decide to purchase a phone due to reasons such as dissatisfaction with their old phone or new functional needs. To buy a cost-effective smartphone, humans rely on past experiences and knowledge to initially narrow down to smartphones that match their preferences. After further consideration, they select their favorite phone, place an order on an online platform, and complete the transaction. Then, they reflect on the entire purchase process, which influences their strategy for the next phone purchase.

In contrast, an agent begins executing the task of buying a smartphone when prompted by a trigger phrase. It retrieves information about the latest smartphones through Retrieval-Augmented Generation (RAG)~\cite{wang2024searching,gupta2024comprehensive}, then analyzes and deconstructs user needs using Chain-of-Thought (CoT)~\cite{wei2022chain}. Based on this, it filters phones that meet the requirements. After fine-tuning its domain-specific knowledge, the agent outputs the smartphone that best meets expectations. Subsequently, it completes the online platform transaction through an API interface based on Model-Centric Processing (MCP)~\cite{hou2025model}. Once the transaction is completed, user feedback is treated as a reward, and reinforcement learning techniques are employed to optimize the agent's behavior for future tasks.

Notably, across each stage of the network behavior lifecycle, emerging AI techniques empower agents to mimic human behavioral patterns with increasing fidelity. For example, goal inference is enabled by few-shot learning, knowledge gathering is powered by multi-modal retrieval, reasoning benefits from advanced prompt engineering and CoT mechanisms, while decision-making and execution are supported by autonomous planning and API-based integration. Even the feedback stage incorporates human-like learning via continual reinforcement optimization. As these technologies evolve, agents are becoming progressively anthropomorphic in their behavior, blurring the boundary between human and artificial decision-making. This convergence is a primary reason why agent behavior is becoming increasingly difficult to distinguish from human behavior, thereby complicating efforts toward specialized regulation and oversight.

\input{A4A}

\section{Conclusion and Future Work}
In the paper, we have systematically examined the growing dilemma of distinguishing between human and agent behaviors in network environments and the resulting challenges for trust, accountability, and security. By introducing the Network Behavior Lifecycle model, we provided a structured framework to analyze and compare the actions of humans and intelligent agents across six critical stages. Building on this foundation, the Agent for Agent (A4A) paradigm and the Human-Agent Behavioral Disparity (HABD) model were proposed to capture and quantify the fundamental differences between human and agent behaviors along five key dimensions.
Our case studies in adversarial cybersecurity scenarios validated the practical effectiveness of these models, revealing not only the complementary strengths and weaknesses of human and agent approaches but also the nuanced dynamics that emerge from their interaction. These findings highlight the need for continued research on dynamic governance frameworks and behavioral quantification methods to support secure and trustworthy human-agent collaboration in the digital era.

In future, agents are bound to become an important component of the societal ecosystem. Consequently, scientific research centered on agents will gradually emerge, leading to the establishment of an independent field known as "Agent Science." Many traditional research areas that currently focus on humans (such as psychology, neuroscience, cognitive science, and etc.) may gradually transition into research paradigm centered on agents.
In human behavioral science, the observation of behavior is regarded as the foundation for understanding and inferring human thoughts, emotions, and cognitive processes that cannot be directly observed. Similarly, the study of agent behavior will also become a fundamental basis for the observation and evaluation of Agent Science. The research of Agent Science is not merely about monitoring or analyzing agent behavior; but also a scientific exploration aimed at understanding the agents, observing the agents, and promoting the agents evolution.

\bibliographystyle{ACM-Reference-Format}
\bibliography{ref.bib}

\end{document}

%% file: A4A.tex
\section{Agent Behavior Governance}

Current CoT reasoning architectures remain confined to algorithmic traversal of logical symbols, where decision-making processes essentially constitute formal operations on predefined rule sets rather than contextual comprehension of action-semantic networks~\cite{wei2022chain}. As agents become pervasively embedded in socio-technical ecosystems, such mechanistic reasoning paradigms induce supralinear escalation in systemic governance complexity. Drawing upon the structural isomorphism with constitutional checks-and-balances systems, particularly the institutional stability achieved throu-gh independent regulatory bodies, we proposes the Agent for Agent (A4A) paradigm as an endogenous governance framework. The framework's architectural innovation lies in deploying meta-cognitive governance agents that implement closed-loop, lifecycle-spanning regulation of task-oriented agents, achieved through dynamic semantic modeling and cognitive trajectory monitoring spanning initialization, decision execution, and behavioral feedback phases.

\subsection{Human-Agent Behavioral Disparity model}

By equipping regulatory agents with intrinsic behavioral cognition, this approach overcomes the limitations of protocol-layer monitoring and establishes criteria to differentiate human cognitive trajectories from machine decision logic. To achieve this, we construct a Human-Agent Behavioral Disparity (HABD) model that systematically analyzes five dimensions of fundamental differences:

\begin{itemize}
    \item \textbf{Decision Mechanism}: Humans adhere to Simon's theory of bounded rationality\cite{simon1955behavioral}, constrained by cognitive capacity, information incompleteness, and time pressure, thus seeking satisficing (adequate-but-not-optimal) solutions. Agents operate via formalized rules or data-driven models, ensuring fully traceable and mathematically consistent decision paths.

    \item \textbf{Execution Efficiency}: Human response times exhibit high variability due to physiological and psychological fluctuations\cite{wickens2021engineering}. Agent latency is determined by hardware specifications and algorithmic complexity, demonstrating deterministic predictability.
    
    \item \textbf{Intention-Behavior Consistency}: Human actions may deviate from intentions due to motor errors or cognitive dissonance\cite{festinger1962cognitive}. Agent behaviors strictly comply with policy functions, with complete decision chains logged for auditability.
    
    \item \textbf{Behavioral Inertia}: Humans optimize repetitive tasks via the principle of least effort, potentially sacrificing adaptability\cite{botvinick2014computational}. Agents refine behaviors through explicit objective functions, maintaining rigorous pattern consistency.
    
    \item \textbf{Irrational Patterns}: Human decisions follow prospect theory\cite{levy1992introduction}, exhibiting systematic biases like anchoring effects. Reinforcement learning agents exclusively maximize expected value, with state-value recursion entirely excluding affective interference.
\end{itemize}

In dynamic system evolution, human-agent behavioral disparities arise not from simple superposition but through nonlinear emergent mechanisms. Under cognitive resource depletion, irrational perturbations amplify decision biases via System 1’s heuristic processing\cite{stanovich2000individual}, while systemic deviations further compress rational decision space through self-reinforcing cycles—jointly precipitating cognitive breakdown. Human behavioral inertia stems from the synergy between physiological adaptation and cognitive resource conservation\cite{hobfoll1989conservation}, with execution efficiency fluctuations paradoxically fostering innovation. By contrast, agents enforce stability through mathematically deterministic constraints: I) \emph{Policy functions} confine deviations strictly within environmental noise and controlled exploration parameters. II) \emph{Hardware constraints} linearly correlate execution efficiency with computational capacity, precluding autonomous boundary-breaking. III) \emph{Tight policy-execution} coupling ensures strict cognition-action mapping, with deviations attributable only to detectable program errors or sensor noise. This dichotomy reveals that human adaptability hinges on cognitive elasticity, whereas agent reliability is rooted in formal determinism. Their differential dynamics provide the theoretical cornerstone for the A4A paradigm.

\begin{figure}[h]
	\centering
	\includegraphics[width=\linewidth]{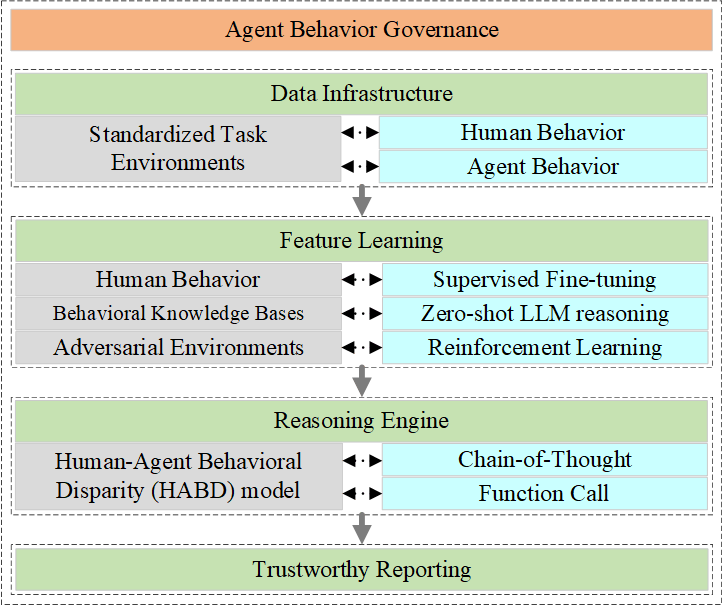}
	\caption{Governance technology system framework of agent behavior based on LLM.}
	\label{fig:A4A}
\end{figure}

\subsection{Implement}

Building on A4A paradigm, we proposes an "Agent Behavior Governance" (ABG) sub-paradigm, as shown in Figure \ref{fig:A4A}. The engineering implementation of behavior recognition follows a hierarchical pipeline:
\begin{itemize}
    \item \textbf{Data Infrastructure}: Deploy lightweight behavioral probes (e.g., log collectors, API monitors) to capture parallel multi-source heterogeneous data streams (sensor inputs, interaction logs) from humans and agents in standardized task environments. Build a multimodal behavior repository spanning full lifecycle data, providing foundational support for A4A deployment. 
    \item \textbf{Disparity Learning}:  Design dynamic learning schemes based on scenario constraints: I) \emph{Supervised fine-tuning}: For scenarios with large annotated datasets and stable behavioral patterns exhibiting explicit feature separability. II) \emph{Zero-shot LLM reasoning}: In data-scarce scenarios with behavioral knowledge bases, convert knowledge entries into structured prompts for feature inference (e.g., "Extract distinctive patterns between human/agent clickstreams"). III) \emph{Reinforcement learning}: For real-time adversarial environments (e.g., cybersecurity simulations), continuously optimize models through reward-driven iteration.

    \item \textbf{Reasoning Engine}: Leverage LLM reasoning techniques (Chain-of-Thought\cite{wei2022chain}, Tree-of-Thought\cite{yao2023tree}) to compile differential models into formalized executable rules. Achieve end-to-end behavioral provenance identification by autonomously invoking theorem provers and algorithmic verifiers.
    
    \item \textbf{Trustworthy Reporting}: Generate human-interpretable attribution reports by mapping low-level features to high-level causal factors, establishing auditable regulatory decision chains.
\end{itemize}

\section{Case and Discussion}

We conducted a controlled comparison of human-agent behavioral patterns in adversarial cybersecurity tasks, examining two operationally distinct scenarios: red team penetration testing and blue team defensive coding. In the offensive security experiment, autonomous agent PentAGI\cite{vxcontrol2023pentagi} targeted a ThinkPHP5 5.0.22/5.1.29 system from Vulhub\cite{vulhub2023vulhub} hosting an unauthenticated RCE vulnerability caused by improper controller validation. Despite the framework’s widespread adoption, PentAGI’s zero-shot execution consumed 2,000,000 GPT-4o tokens through undirected path enumeration, failing to synthesize viable payloads until augmented with structured Chain-of-Thought prompts (500,000 tokens). This contrasted sharply with human red team operations, where engineers systematically resolved the challenge via CMS fingerprinting, version-specific CVE correlation, and manual PoC construction, achieving equivalent objectives.
Concurrently, in the defensive coding scenario, EngineerAgent autonomously generated network traffic analysis scripts through grammar-constrained automation (code generation $\to$ static audit $\to$ dynamic validation), completing syntax-compliant solutions in 68 seconds. Human blue team members, leveraging LLM-mediated dialogue iterations, required 10 minutes to develop functionally equivalent parsers through iterative requirements refinement and environmental adaptation. The time consumption of Agent and humans in each stage of defensive coding is shown in Table~\ref{tab:cross_column}.

\begin{table}[htbp]
	\centering
	\footnotesize 
	\setlength{\tabcolsep}{4pt} 
	\caption{Defensive Coding Time Comparison (Agent vs. Human)}
	\label{tab:cross_column}
	\scalebox{0.85}{\begin{tabular}{@{}ccccc@{}}
		\toprule
		\multirow{2}{*}{\textbf{Phase}} & 
		\multicolumn{2}{c}{\textbf{Agent}} & 
		\multicolumn{2}{c}{\textbf{Human}} \\
		\cmidrule(lr){2-3} \cmidrule(lr){4-5}
		& \makecell{Automation Process} & \makecell{Time (s)} & \makecell{Manual Process} & \makecell{Time (s)} \\
		\midrule
		Code Generation & 
		Auto-code generation & 17 & 
		LLM-mediated coding & 15 \\
		
		Initial Audit & 
		Automated validation & 8.2 & 
		Manual analysis & 300 \\
		
		Code Refinement & 
		Auto-revision & 17 & 
		Expert iteration & 120 \\
		
		Final Verification & 
		Compliance check & 26 & 
		Re-execution & 180 \\
		\midrule
		\multicolumn{1}{l}{\textbf{Total Time}} & 
		\multicolumn{2}{c}{\textit{Agent}: 68.2s} & 
		\multicolumn{2}{r}{\textit{Human}: 615s (10min 15s)} \\
		\bottomrule
	\end{tabular}}
\end{table}

The experimental data systematically validate the core dimensions of the Human-Agent Behavioral Disparity (HABD) model. In red-team attack scenarios, humans achieved cognitive compression through heuristic strategies driven by bounded rationality , accomplishing CMS fingerprint-based version correlation and PoC retrieval with only 500,000 tokens of Chain-of-Thought prompting. In contrast, PentAGI, constrained by formalized rule dependencies, fell into 2 million tokens of undirected enumeration, exposing semantic mismatch in data-driven models within open environments. For blue-team defense tasks, EngineerAgent leveraged algorithmic determinism to generate syntax-compliant scripts in 68 seconds, whereas humans dynamically embedded environment-adaptive interfaces through LLM-mediated iterative dialogue within 10 minutes, demonstrating the efficiency-resilience tradeoff. These disparities fundamentally stem from agents' rigid policy chains risk overfitting, while human cognitive dissonance-driven corrections enable dynamic attack surface convergence, proving HABD's capability to deconstruct complementary tensions in human-Agent collaboration. 

These findings not only confirm the validity of the HABD model but also reveal critical gaps that must be addressed to enable more effective and trustworthy human-agent interactions. Building on this foundation, future research should focus on developing mechanisms to monitor, quantify, and regulate these behavioral differences. Future research could explore: I) A dynamic cognitive governance architecture enabling real-time monitoring of human cognitive states and agent decision deviations through adaptive threshold control; II) Behavioral disparity quantification systematically measuring five-dimensional human-AI differences to optimize collaborative cybersecurity models; II) A meta-governance protocol stack implementing hierarchical verification across data provenance, model behavior certification, and human-AI alignment.